\documentclass[letterpaper, 10 pt, conference]{ieeeconf}  
\usepackage{amssymb}
\IEEEoverridecommandlockouts                              

\usepackage{graphicx}
\usepackage{url}
\usepackage{multicol}
\usepackage{graphicx}
\usepackage{color}
\usepackage[pagebackref=true,breaklinks=true,colorlinks,bookmarks=false]{hyperref}
\usepackage[export]{adjustbox}
\usepackage{subcaption}
\usepackage{wrapfig}

\usepackage{multicol}
\usepackage{booktabs}
\usepackage{algorithm}
\usepackage[noend]{algpseudocode}
\usepackage{float}
\usepackage{amsmath, amssymb}
\usepackage{amsfonts}
\usepackage[skip=2pt,font=small,labelfont=bf]{caption}
\usepackage{bm}
\usepackage{epstopdf}

\usepackage{todonotes}

\definecolor{purple}{RGB}{210, 0, 210}
\definecolor{dgreen}{RGB}{34, 139, 32}

\definecolor{international_orange}{RGB}{240, 74, 0}

\usepackage{cite} %

\newcommand{\reals}{\mathbb{R}}
\renewcommand{\vec}{\mathbf}
\newcommand{\ggn}{\emph{DefGoalNet}}
\newcommand{\DeformerNet}{\emph{DeformerNet}}
\newcommand{\pcloud}{\mathcal{P}}
\newcommand{\curpcloud}{\pcloud_\mathrm{c}}
\newcommand{\goalpcloud}{\pcloud_\mathrm{g}}

\newcommand{\object}{\mathcal{O}}
\newcommand{\objectgoal}{\object_\mathrm{g}}
\newcommand{\policy}{\pi}
\newcommand{\sspolicy}{\pi_\mathrm{s}}
\newcommand{\action}{\mathcal{A}}

\newcommand{\manippoint}{\vec{p}_\mathrm{m}}
\newcommand{\encoder}{g}

\newcommand{\featcur}{\psi_\mathrm{c}}

\newcommand{\featconcat}{\psi_\mathrm{f}}
\newcommand{\feattask}{\psi_{T}}

\newcommand{\demodata}{\mathcal{D}}
\newcommand{\task}{T}
\newcommand{\taskpcloud}{\pcloud_\task}
\newcommand{\trajectory}{\tau}
\newcommand{\numpoints}{N}
\newcommand{\decoder}{d}
\newcommand{\se}{\mathcal{SE}}

\title{DefGoalNet: Contextual Goal Learning from Demonstrations For Deformable Object Manipulation}
    
\begin{document}
\author{Bao Thach\(^{1,\mathsection}\), Tanner Watts\(^{1,\mathsection}\), Shing-Hei Ho\(^1\), Tucker Hermans\(^{1,2}\), Alan Kuntz\(^1\)
\thanks{$\mathsection$ These authors contributed equally. $^{1}$Robotics Center and Kahlert School of Computing, University of Utah, Salt Lake City, UT 84112, USA; $^{2}$NVIDIA Corporation, Seattle, WA, USA; This work was supported in part by NSF Awards \#2024778 and \#2133027. T. Watts and S. H. Ho were also supported in part by funding from the Undergraduate Research Opportunities Program at the University of Utah.
{\tt\{bao.thach, tanner.watts, shinghei.ho, tucker.hermans, alan.kuntz\}@utah.edu}
}}
\input{intro_figure}
\maketitle

\urlstyle{same}

\begin{abstract}

Shape servoing, a robotic task dedicated to controlling objects to desired goal shapes, is a promising approach to deformable object manipulation.
An issue arises, however, with the reliance on the specification of a goal shape.
This goal has been obtained either by a laborious domain knowledge engineering process or by manually manipulating the object into the desired shape and capturing the goal shape at that specific moment, both of which are impractical in various robotic applications.
In this paper, we solve this problem by developing a novel neural network \ggn{}, which learns deformable object goal shapes directly from a small number of human demonstrations.
We demonstrate our method's effectiveness on various robotic tasks, both in simulation and on a physical robot. 
Notably, in the surgical retraction task, even when trained with as few as 10 demonstrations, our method achieves a median success percentage of nearly 90\%.   
These results mark a substantial advancement in enabling shape servoing methods to bring deformable object manipulation closer to practical real-world applications.
\end{abstract}
\vspace{-5pt}

\section{Introduction}\label{sec:intro}
Deformable object manipulation defines a fundamental challenge in robotic manipulation due to its wide-ranging applications~\cite{Sanchez2018robotic, zhu2022challenges}.
Unlike rigid objects, deformable materials such as clothing, fabrics, soft tissues, or food items have intricate dynamics and infinite degrees of freedom.
This complexity necessitates innovative techniques to enable robots to manipulate deformable objects effectively.
Whether in healthcare (surgical robots for delicate tissue manipulation), manufacturing (handling soft materials like textiles), or homes (folding laundry), enabling effective deformable object manipulation empowers robots to perform a broader range of tasks, increasing their utility.

Shape servoing, a robotic task dedicated to controlling objects to desired goal shapes, has recently garnered great attention from the deformable object manipulation community~\cite{thach2023deformernet,thach2022learning,thach2021deformernet,Navarro-Alarcon2013b, Navarro-Alarcon2014, Navarro-Alarcon2016, qi2022model, alambeigi2018robust, alambeigi2018autonomous,Navarro-Alarcon2017,Hu20193-D, Qi2019Contour,zhu2021vision}. 
Our prior work in this area, \DeformerNet{}~\cite{thach2023deformernet}, leverages point clouds as the state representation for deformable objects.
\DeformerNet{} defines a neural network that takes the object's current and goal point clouds as inputs and computes the desired robot action to drive the object toward the target shape.
However, a significant weakness of \DeformerNet{} and other shape servoing methods is the requirement to explicitly define goal shapes, e.g., as a point cloud.
Goals have previously been obtained either by laborious domain knowledge engineering or by manual manipulation of an object into the desired shape and capturing the shape at that specific moment---approaches that often prove impractical in various robotic applications. 

We address this critical challenge by introducing \ggn{}, a novel neural network capable of autonomously learning deformable object goal shapes from human task demonstrations.
Our method fills the vital gap in \DeformerNet{} by supplying it with a goal point cloud predicted from additional sensory information encoding the task context.
We denote this as the \textit{contextual point cloud}. The contextual point cloud provides crucial features for defining the success of a specific task.
The design choice for the contextual point cloud varies by task but will generally include other task-relevant objects or environment components.
When integrating \ggn{} and \DeformerNet{} into a unified pipeline, we transform the robot policy formulation, making it reliant solely on practically accessible parameters: the current point cloud of the deformable object and an additional point cloud encoding contextual information specific to the current task.

In our new learning-from-demonstration shape servoing pipeline, we first train \ggn{} on a set of demonstration trajectories.
This enables it to predict goal shapes that vary as task context changes. 
At runtime, our neural network reasons over the current deformable object point cloud and the contextual point cloud, generating a goal point cloud corresponding to a successful task outcome under that context.
This output from \ggn{} then becomes the input to \DeformerNet{}, which computes the desired robot end-effector action to drive the deformable object toward the goal shape, accomplishing the task in a closed-loop manner (see Fig.~1).
This pipeline decouples goal generation from the control policy. This in turn enables learning a robust policy across a large set of related data (e.g., from simulation) while learning goals from potentially few samples of the target task.

We evaluate our method with experiments on surgery-inspired robotic tasks. 
We first conduct experiments in simulation, with simulated demonstrations, on two common surgical sub-tasks: retraction and tissue wrapping.
We then perform a zero-shot sim-to-real transfer experiment with a physical robot, using the model entirely trained on simulated data.
Finally, we train \ggn{} with a small set of real human demonstrations and evaluate our method on a physical mock surgical retraction task.
In all experiments, we find \ggn{} achieves excellent performance.
Notably, on a surgical retraction task, our method achieves high performance when trained on as few as 10 demonstrations. %

While a small demonstration dataset can still lead to the successful completion of the task, our experiments show that increasing the size of the demonstration dataset results in more realistic and easily interpretable goal shape prediction.
Visualizing these high-quality goal shapes as an intermediate step before executing the robot policy could enhance safety and transparency in robot learning from demonstration. 

Our paper is the first effort to tackle the challenge of learning goal shape specification in shape servoing tasks from demonstration.
The experimental outcomes we have achieved represent a significant stride toward making shape servoing more applicable to real-world robotic applications.
We publish all code and data at: \\
\url{https://sites.google.com/view/defgoalnet/home}.

\section{Related Work}\label{sec:related_work}
Machine learning has endowed robots with the capability to adeptly manipulate rigid objects by harnessing complex, high-dimensional sensor data, such as point clouds~\cite{lu2020multifingered, mousavian20196, murali20206, deng2020self, lu2020multi, van2020learning}. The integration of neural networks has revolutionized the resolution of challenging robotic tasks, such as shape completion~\cite{van2020learning}, pose estimation~\cite{lu2020multi}, and grasping~\cite{lu2020multifingered, mousavian20196, lu2020multi, van2020learning}. Even tasks demanding long-horizon planning and a range of skills, such as the comprehensive removal of all food items from a table~\cite{brohan2022rt}, have found successful solutions by deploying cutting-edge learning tools. Inspired by these advancements, we examine a learning-based method to address 3D deformable object shape control.

Shape servoing historically has been tackled by mostly learning-free approaches~\cite{navarro2013visually, alambeigi2018autonomous,navarro2016automatic,qi2022model,alambeigi2018robust,shetab2022lattice}.
These methods often represent the object as a set of hand-picked feature points, thus struggling to generalize to unseen objects and being too vulnerable to sensor noise.
Shetab-Bushehri \emph{et al.}~\cite{shetab2022lattice} leverage a 3D lattice to describe deformable objects enabling accurate 3D shape control. Nevertheless, an assumption of feature correspondence limits its application. 

Hu \emph{et al.}~\cite{Hu20193-D} use the fast point feature histogram (FPFH)~\cite{Rusu2010VFH} of deformable objects as the state representation for shape control learning. However, Thach \emph{et al.}~\cite{thach2022learning} showed that the FPFH fails to capture the complex dynamics of 3D deformable objects.
\DeformerNet{}~\cite{thach2022learning,thach2023deformernet} defines the current state-of-the-art to 3D shape servoing, operating effectively both in simulation and physical-robot experiments. 
It offers a neural network that inputs the object's current and goal point clouds and computes the desired robot action to drive the object toward the target shape. 
However, a significant weakness of \DeformerNet{} and other shape servoing methods is the requirement to define goal shapes explicitly.

While point cloud generative networks show strong performance~\cite{shu20193d, achlioptas2018learning, arshad2020progressive,pumarola2020c}, their application in robotics is somewhat limited. In terms of goal generation for robotics, Waveren \emph{et al.}~\cite{van2022large} introduce a generative model that can render a goal image for a rearrangement task, given natural language instructions.
However, this method does not reason about changing deformable object geometries.

Various learning-based methods have been examined for robotic automation of surgical procedures, such as cutting~\cite{Thananjeyan2017_ICRA, Murali2015_ICRA}, suturing~\cite{vandenberg2010_ICRA,Chiu2021Bimanual}, 
retracting tissues~\cite{Attanasio2020_RAL,pore2021safe},
navigating surgical tools~\cite{kim2020autonomously}, and tissue tracking ~\cite{Lu2021Super,lin2022semantic}.

\section{Problem Formulation}\label{sec:problem}
We address the problem of orchestrating robotic manipulation of a 3D deformable object to achieve a specific task $\task$. 
The term \textit{3D}, herein interchangeably referred to as \textit{volumetric}, indicates that no single dimension in the object significantly surpasses the other two in scale~\cite{Sanchez2018robotic}.

We define the 3D volumetric object to be manipulated as $\object \subset \reals^3$. $\object$ can undergo dynamic changes during robotic manipulation.
Due to the inherent limitations in directly sensing the entirety of $\object$, we instead work with a partial-view point cloud $\pcloud \subset \object$, which encompasses a subset of points on the object's surface. 
We represent the current point cloud of the object as $\curpcloud$.

The robot can reshape the object by grasping it at a pair of manipulation points $\{{\manippoint}_i\}_{i=1:2}$ and subsequently moving its end-effectors. 
We define robot actions $\action$ as a pair of homogeneous transformation matrices: $\action \in \se (3) \times \se (3)$, representing the desired change of end-effector poses.
Note that for tasks that do not require bimanual manipulation, we only need a single manipulation point and a single transformation matrix for the action.

We frame our problem as a contextual learning problem wherein there is observable context present at the start of the task execution, which, when combined with the observable state of the deformable object, dictates how the task should be performed.
We encode the task-specific context as a contextual point cloud $\taskpcloud$. This point cloud does not belong to the object volume but provides crucial features for the task's success.
It may include task-relevant objects or features in the surrounding environment.
The design choice for the contextual point cloud varies from task to task and will be elaborated on in Sec.~\ref{sec:experiments}.
For instance, for a surgical robot tasked with lifting a deformable tissue layer off a kidney, $ \taskpcloud$ could include the partial view of the kidney observable at the start of the task execution.

For a given task $\task$, we assume an associated dataset, $\demodata$, which consists of a set of demonstration trajectories that accomplish the task under various contexts. We define each trajectory $\trajectory$ as an ordered sequence of $M$ waypoints $(\pcloud_1, \pcloud_2, \ldots, \pcloud_{M})$, where each $\pcloud_i$ is a point cloud observation of the deformable object at the associated time.
The problem then becomes to learn a policy to autonomously perform $\task$ in new, unseen contexts. To learn this, the robot has the available training data of associated contexts, initial object states, and demonstration trajectories.

\section{Method}\label{sec:method}
Our formulation assumes that a goal shape $\objectgoal$ exists, whereby if the object reaches this particular configuration, the task succeeds.
Leveraging this intuition, we decompose the initial complex problem into two distinct sub-problems. 

The first sub-problem examines contextual goal learning. Here, we aim to predict a goal point cloud $\goalpcloud$ corresponding to a successful task outcome, based on the current state and the context: $\goalpcloud = \Phi(\curpcloud, \taskpcloud)$.
We can train a task specific goal generation model using the demonstration dataset $\demodata$ associated with task \(\task\).
At runtime, our neural network reasons over the current point cloud observations and generates a goal point cloud that corresponds to a successful task outcome under that context.

Given this point cloud, we can examine the second question of goal-conditioned shape control. 
We address this by learning a policy that maps the current point cloud, goal point cloud, and manipulation point to a robot action: $\policy(\curpcloud, \goalpcloud, \manippoint) = \action$.
By applying this policy repeatedly, the robot gradually transforms the object towards its goal shape, ultimately accomplishing the task.
The robot selects manipulation points using the \textit{dense predictor} network from~\cite{thach2023deformernet}.

Note that \DeformerNet{}~\cite{thach2023deformernet} presents an effective solution to the second sub-problem. Therefore, the remainder of this section primarily focuses on the contextual goal learning problem. 
We first provide details of the proposed architecture for \ggn{}. We then explain how we train the model. We conclude the section by describing how we integrate the goal generation network with \DeformerNet{}.

\subsection{\ggn{} Architecture Details}
We adopt an encoder-decoder architecture for \ggn{}. Our neural network feeds the inputs $\curpcloud$ and $\taskpcloud$ into two identical PointConv~\cite{PointConv2019} encoder channels $\encoder$, generating two feature vectors $\featcur = \encoder(\curpcloud)$ and $\feattask = \encoder(\taskpcloud)$. By concatenating them together, we obtain the final feature vector \(\featconcat = \featcur\bigodot\feattask\). This feature vector is fed into a decoder $\decoder$, which consists of a series of fully connected layers and eventually outputs a 1D vector with $3\times\numpoints$ elements, where $\numpoints$ is the desired number of points in the goal point cloud. The 1D vector is then reshaped to construct a point cloud of shape $3\times\numpoints$.
The composite goal generator thus takes the form: $\goalpcloud = \Phi(\curpcloud, \taskpcloud) = \decoder(\encoder(\curpcloud)\bigodot\encoder(\taskpcloud))$.

Prior to training, we employ farthest point sampling~\cite{Qi2017PointNet} to downsample both $\curpcloud$ and $\taskpcloud$ to $\numpoints$ points.
In all our experiments, we choose $\numpoints$ to be 512. 

\begin{figure}[ht]
    \centering
    \includegraphics[width=0.9\linewidth, height=14cm]{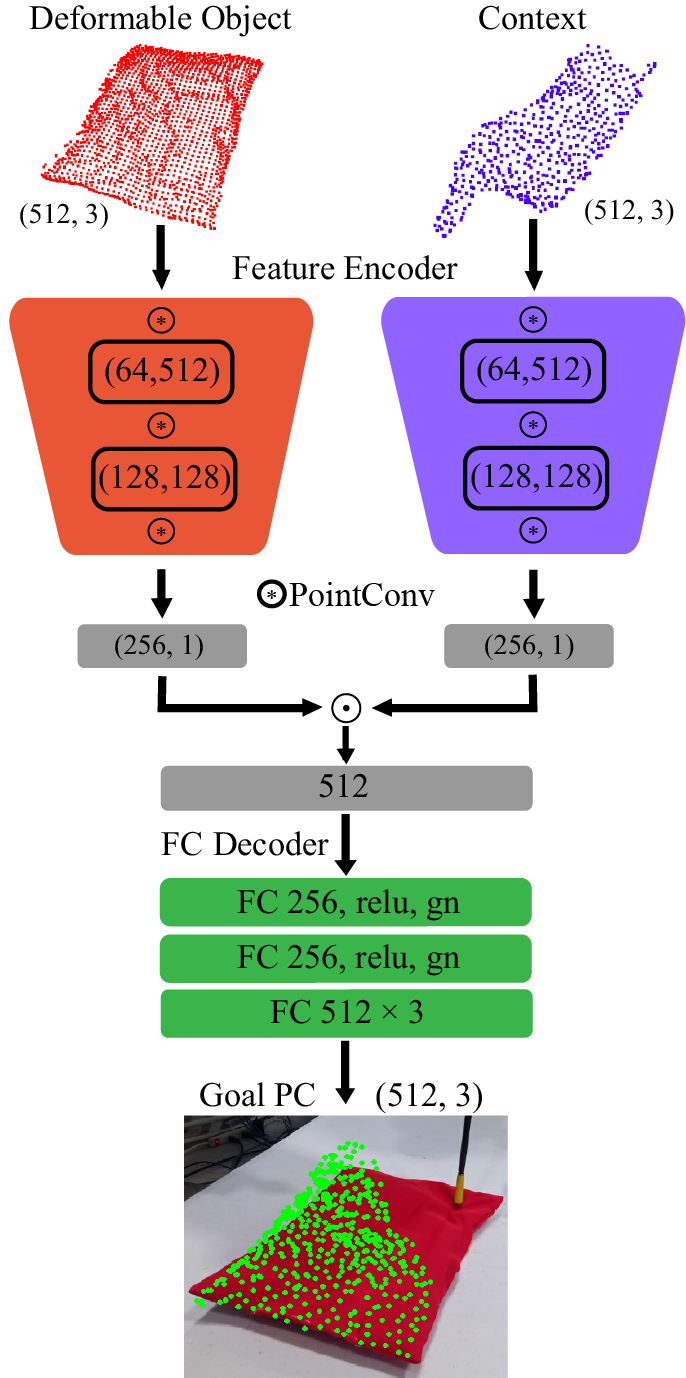}
    \caption{\ggn{} architecture, comprising PointConv-based feature encoders and a fully-connected decoder.}
    \label{fig:goalgennet_architecture}
    \vspace{-16pt}
\end{figure}

Figure~\ref{fig:goalgennet_architecture} provides a comprehensive overview of the \ggn{} architecture. We design the encoder to have three consecutive PointConv~\cite{PointConv2019} layers. These layers progressively output point clouds of dimensions $64\times512$, $128\times128$, culminating in a 256-dimensional feature vector. The decoder architecture encompasses a sequence of fully-connected layers with hidden layer sizes of $256, 256$, and $3\times\numpoints$.

\subsection{\ggn{} Training Procedure}
Training \ggn{} follows a straightforward, supervised-learning approach.
Given a demonstration trajectory, we apply a segmentation mask over the raw point cloud observations to obtain points that belong to the object.  
We set the object point cloud at the beginning of the trajectory as $\curpcloud$ and the terminal object point cloud as \(\goalpcloud\). 
For the remaining points that do not belong to the object volume, we select a subset of task-relevant points and set them as the context $\taskpcloud$. The design choice of what points to be included in $\taskpcloud$ varies from task to task. We elaborate on our specific choices in Sec.~\ref{sec:experiments}.

To effectively capture the complex goal point clouds of deformable objects, we adopt a loss function that combines Chamfer distance and earth mover's distance, both widely recognized point cloud distance metrics~\cite{lin2018learning, wang2022learning}.
Chamfer distance measures the average distance of each point in one set to the nearest point in the other set:
\[C(\vec{p}_a, \vec{p}_b) = \sum_{x\in \vec{p}_a}\min_{y\in \vec{p}_b} ||x - y||^2 + \sum_{y\in \vec{p}_b}\min_{x\in \vec{p}_a} ||x - y||^2 .\]
In contrast, earth mover's distance quantifies the dissimilarity between two point distributions:
\[E(\vec{p}_a, \vec{p}_b) = \min_{\xi: \vec{p}_a \rightarrow \vec{p}_b}\sum_{x\in \vec{p}_a}||x - \xi(x)||_2,\]
The training loss, a linear combination of these two distances, measures the disparity between predicted and ground-truth point clouds.
We train \ggn{} end-to-end using the Adam solver~\cite{kingma2014adam} on a single RTX 3090 GPU.

\subsection{Integration with \DeformerNet{}}
\DeformerNet{} provides an effective policy for deformable object manipulation: \(\sspolicy(\curpcloud, \goalpcloud, \manippoint) = \action \). The shape servoing policy takes as inputs a manipulation point along with the current and goal object point clouds. 
An issue arises, however, with the reliance on the specification of a goal shape, which is impossible to obtain in some applications.

Herein lies the significance of \ggn{}, which generates a goal point cloud $\goalpcloud = \Phi(\curpcloud, \taskpcloud)$. This predicted goal point cloud becomes the input for \DeformerNet{}. As a result, the robot policy within \DeformerNet{} no longer hinges on the enigmatic $\goalpcloud$, but is instead formulated as a function of practically obtainable parameters, namely $\curpcloud$ and $\taskpcloud$:
\[\sspolicy(\curpcloud, \goalpcloud, \manippoint) = \sspolicy(\curpcloud, \Phi(\curpcloud, \taskpcloud), \manippoint)= \action.\]
Running this policy in a closed-loop fashion, the robot will accomplish the task.
Prior to manipulation, given the learned $\goalpcloud$, the \textit{dense predictor} network of \DeformerNet{} can select the manipulation points on the object that the robot should grasp.
Note that we leverage the \DeformerNet{} and \textit{dense predictor} models trained in the original paper\cite{thach2023deformernet} without any fine-tuning.
This demonstrates an advantage of our approach, enabling independent training of goal prediction and deformable shape control.

\section{Experiments and results}\label{sec:experiments}
We assess the performance of our method in simulation and on a real-world robotic setup. In simulation, we employ a model of the patient-side manipulator from the da Vinci research kit surgical robot%
\cite{Kazanzides2014_ICRA_DVRK} using the Isaac Gym platform~\cite{Liang2018GPU}. We test on a Baxter robot equipped with a laparoscopic tool and an Azure Kinect camera for capturing point clouds for our real-world experiments.

We evaluate our method on two simulation-based robotic tasks: surgical retraction and tissue wrapping. We then conduct a zero-shot sim-to-real transfer experiment with a physical robot, using the model trained entirely on simulated data. Finally, we experiment with a real human demonstration dataset on a physical mock surgical retraction task.

\begin{figure}[ht]
    \centering
    \includegraphics[width=0.8\linewidth]{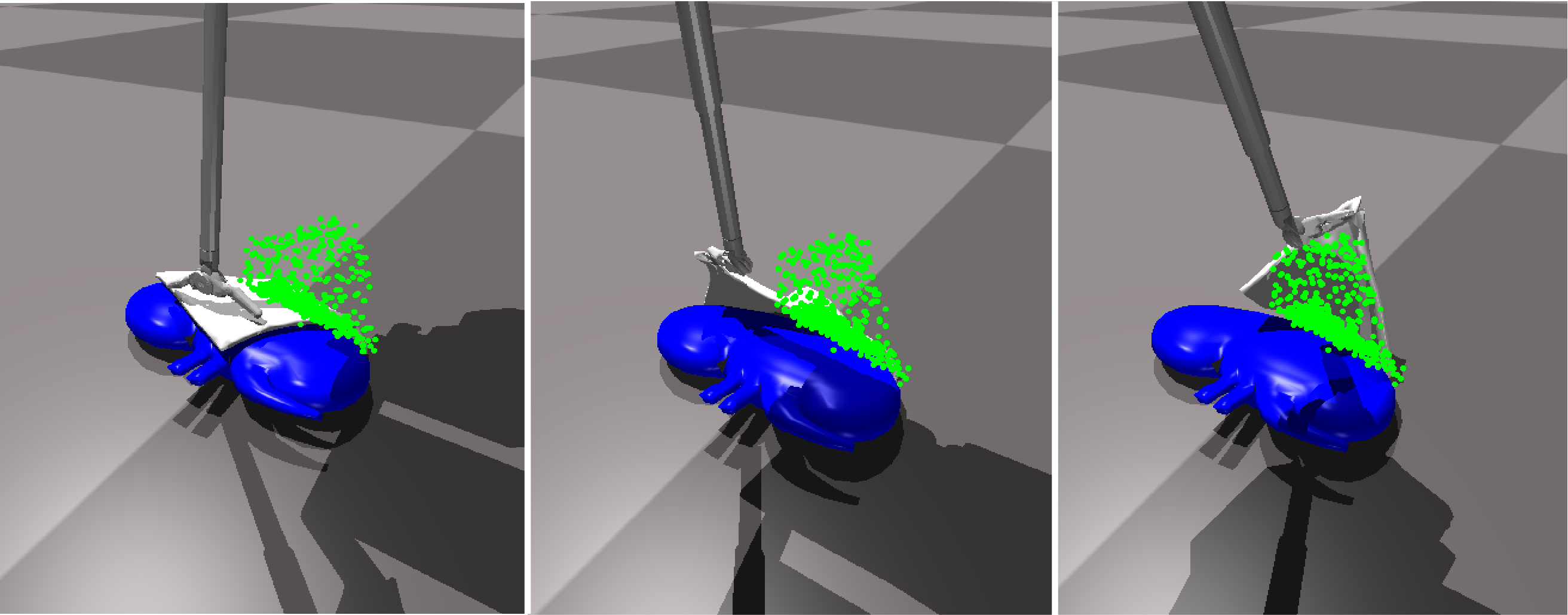}
    \caption{Sample manipulation sequence on the surgical retraction task, in simulation. The green point cloud visualizes the goal shape generated by \ggn{}.}
    \label{fig:sequence_retraction_kidney_sim}
    \vspace{-10pt}
\end{figure}

\subsection{Demonstration data collection}
We adopt the same procedure for all experiments in this paper to collect the training data for \ggn{}. First, we collect $M$ demonstration trajectories that accomplish the task. These trajectories can be executed by scripted robot actions (as in our simulation experiments) or real human actions (as in our physical robot experiments). We record each trajectory's initial and terminal object point clouds and save them as current and goal point clouds ($\curpcloud$ and $\goalpcloud$), respectively, for training. We also record the contextual point cloud $\taskpcloud$. We detail the design choice for $\taskpcloud$ in the following sections after formally introducing each task. 
Finally, we construct input-output pairs for \ggn{} training. The input is ($\curpcloud, \taskpcloud$), and the output is $\goalpcloud$.

\begin{figure}[ht]
    \centering
    \includegraphics[width=0.95\linewidth, height=4.5cm]{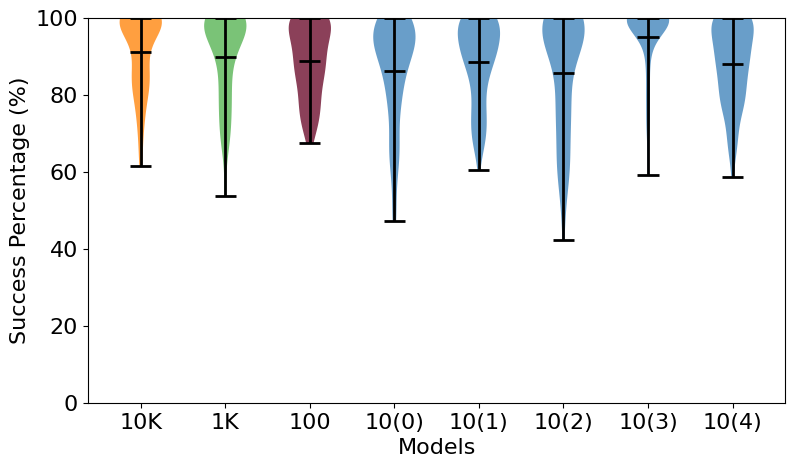}  
    \caption{\textbf{Simulated retraction results} - Success percentage on the test set across multiple training dataset sizes. From left to right: \ggn{} trained with 10000, 1000, 100, and 10 demonstrations. We train the 10-demonstration model 5 times with 5 different sets of demonstrations, which are labeled 10(0-4).}     \label{fig:percentage_plot_retraction_kidney_sim}
    \vspace{-10pt}
\end{figure}
\subsection{Simulation Experiments}
\subsubsection{Surgical Retraction}\label{sec:retraction_exp_sim}
Surgical retraction plays a pivotal role in nephrectomies, a procedure during which an adipose tissue layer needs to be retracted off of a kidney~\cite{ashrafi2020minimally}. Here, we designed a simulated robotic task to emulate this surgical procedure, as depicted in Figure~\ref{fig:sequence_retraction_kidney_sim}. The robot is assigned to lift the tissue to reveal the kidney beneath it.

\begin{figure}[th]
    \centering
    \includegraphics[width=1\linewidth, height=8.5cm]{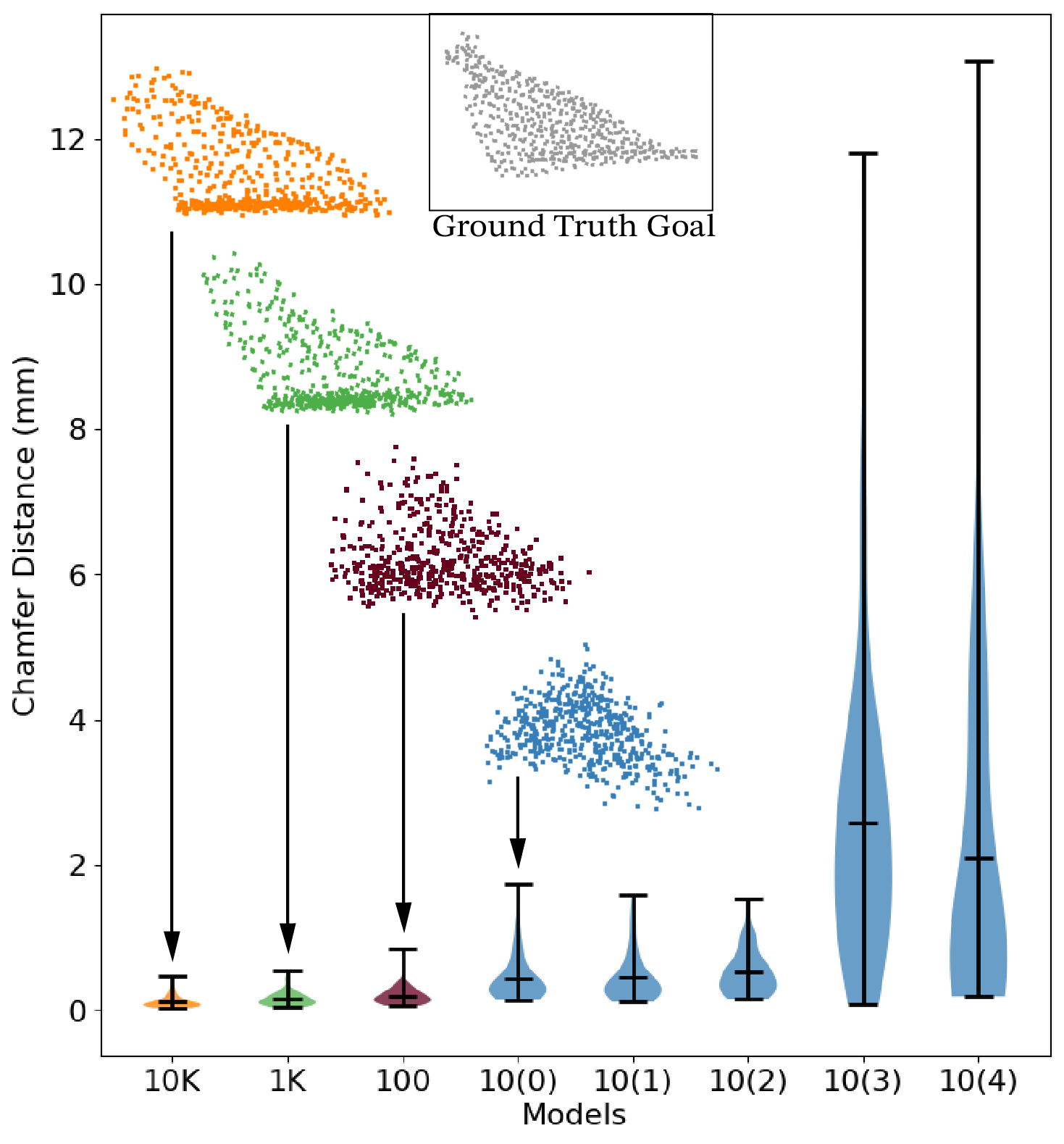}
    \caption{\textbf{Simulated retraction results} - Chamfer distance between predicted and ground truth goal point clouds on the test set across multiple training dataset sizes. Example predicted goals are visualized on top of each violin plot.}    \label{fig:chamfer_plot_retraction_kidney_sim}
    \vspace{-20pt}
\end{figure}
To train \ggn{} on this task, we create a dataset of different kidneys whose sizes are sampled from the distribution of typical, adult human kidneys~\cite{glodny2008normal}, along with different box-like tissue layers whose geometries are sampled from a uniform distribution.
We also randomize the kidney and tissue poses. The contextual point cloud $\taskpcloud$ for this task is the partial-view point cloud of the part of the kidney unoccluded by the adipose layer.
The demonstration trajectories are generated using scripted robot actions.
We set a target plane that bisects the kidney into two halves along its longitudinal dimension. 
The robot demonstrator grasps the tissue and retracts it until the entire tissue layer passes beyond the target plane.

We examine the influence of dataset size on the performance of \ggn{} by training separate models on 10, 100, 1000, and 10000 demonstrations.
The 10000 demonstration set is obtained by running the robot demonstrator on 100 different sampled kidneys, each with 100 different sampled tissue layers, all with randomly sampled poses. The smaller sets are selected uniformly at random from the largest set.
We train the 10 demonstration model 5 times with 5 different sets of demonstrations to examine performance variance at small data scales.
We evaluate our method on a test set comprising 100 unseen configurations (kidney size and pose, and tissue size and pose). A representative manipulation sequence is visualized in Fig.~\ref{fig:sequence_retraction_kidney_sim}.
Two evaluation metrics quantify the performance of our method.

The first and most important metric directly measures the \textit{success rate}. On each test configuration, we run the \DeformerNet{} policy on the predicted goal generated by \ggn{} and record the final object point cloud. 
We leverage the robot demonstrator's target plane to assess how good each retraction trajectory is. We count the percentage of points in the final point cloud that successfully pass through the target plane, which we call \textit{success percentage}.
As visualized in Fig.~\ref{fig:percentage_plot_retraction_kidney_sim}, even with as few as 10 demonstrations, our method can still achieve a median success percentage of nearly 90\%.

The second metric is the Chamfer distance between the predicted and ground truth goal point clouds. The Chamfer results and example predicted goals are visualized in Fig.~\ref{fig:chamfer_plot_retraction_kidney_sim}.
Unsurprisingly, as we increase the demonstration dataset size, the generated goals become more realistic and easily interpretable; however, the structure of the goal is visible even from 10 demonstrations.

\subsubsection{Tissue Wrapping}
To showcase the breadth of deformable manipulation tasks $\ggn$ can handle, we conduct further experiments on a tissue wrapping task inspired by surgical procedures such as aortic stent placement. This entails the cooperation of two robotic arms to encase a thin layer of tissue around a cylindrical tube, with the goal of maximizing tissue coverage on the tube's surface. 
The contextual point cloud $\taskpcloud$ for this task is the partial-view point cloud of the cylindrical tube.

\begin{figure}[ht]
    \centering
    
    \includegraphics[width=0.325\linewidth]{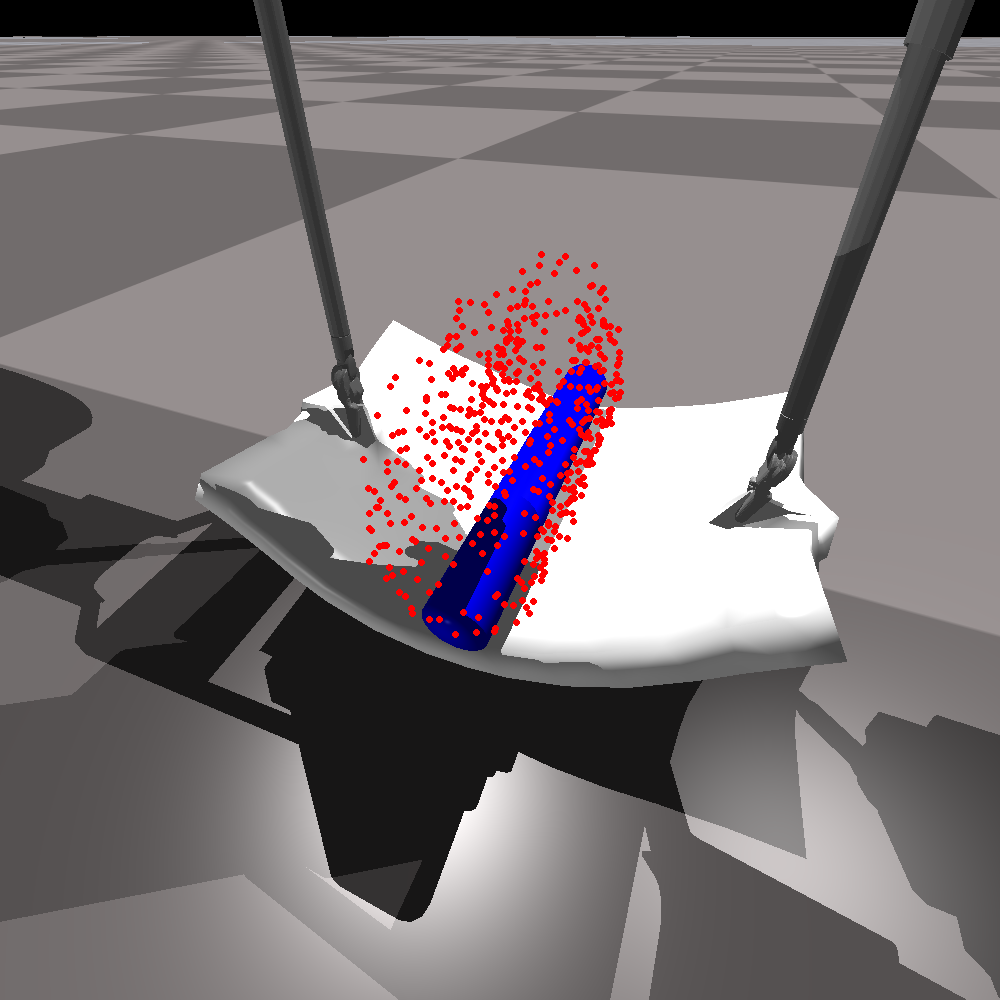} 
    \includegraphics[width=0.325\linewidth]{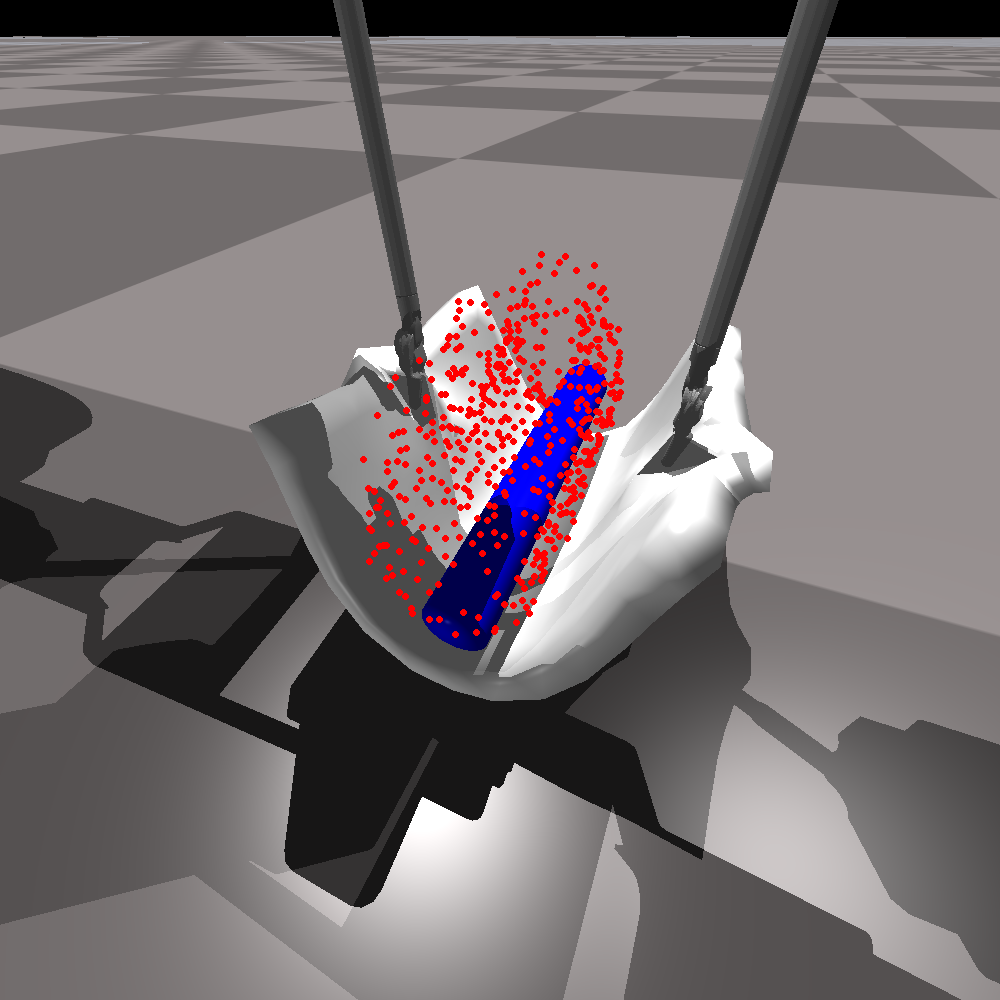}    
    \includegraphics[width=0.325\linewidth]{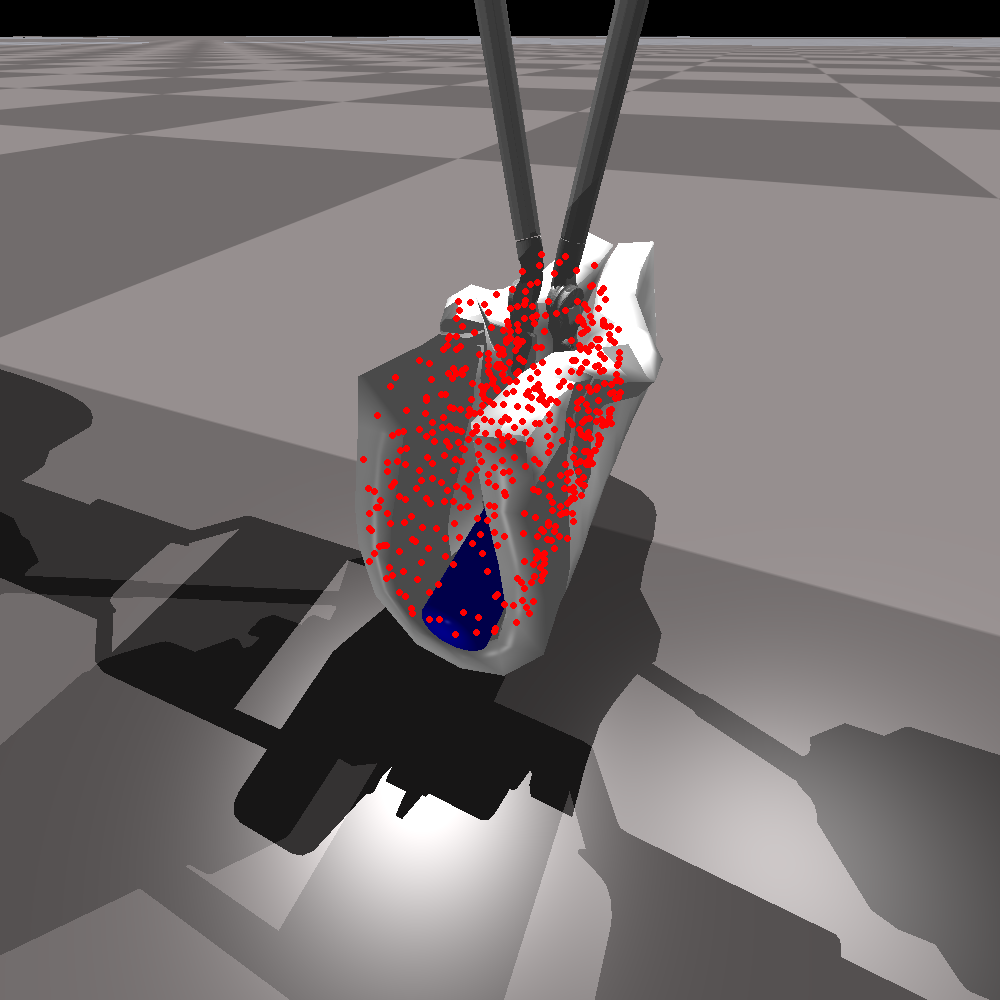} 
 
    \caption{Sample manipulation sequence on the tissue wrapping task, in simulation. The red point cloud visualizes the goal shape generated by \ggn{}.}
    
    \label{fig:sequence_tissue_wrapping_sim}
    \vspace{-20pt}

\end{figure}

We train \ggn{} on a varying number of demonstrations: 10, 100, and 1000. For the case of 10 demonstrations, we also train 5 times with different random seeds to validate the robustness of the approach. A representative manipulation sequence is visualized in Fig.~\ref{fig:sequence_tissue_wrapping_sim}. 

\begin{figure}[ht]
    \centering
    \includegraphics[width=1\linewidth]{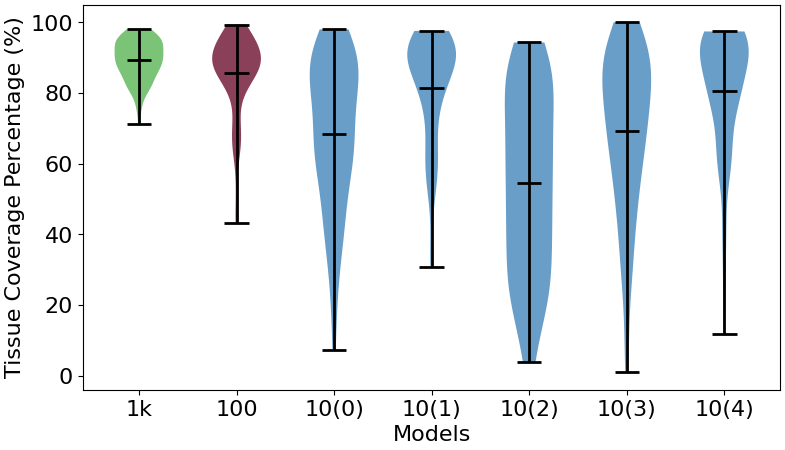}
    \caption{\textbf{Simulated tissue wrapping} - Tissue coverage percentage.} \label{fig:chamfer_plot_coverage_percentage_sim}
    \vspace{-10pt}
\end{figure}

We compute the tissue coverage percentage to quantify task success, developed in~\cite{thach2023deformernet}. This measures the percentage of tube surface area being wrapped by the tissue. 
We run \DeformerNet{} on the test set with goals generated by \ggn{} and visualize the results in Fig.~\ref{fig:chamfer_plot_coverage_percentage_sim}. At a dataset size of 100, our method starts achieving competitive performance with a median coverage percentage of almost 90\%.

\begin{figure}[ht]
    \centering
    \includegraphics[width=1\linewidth]{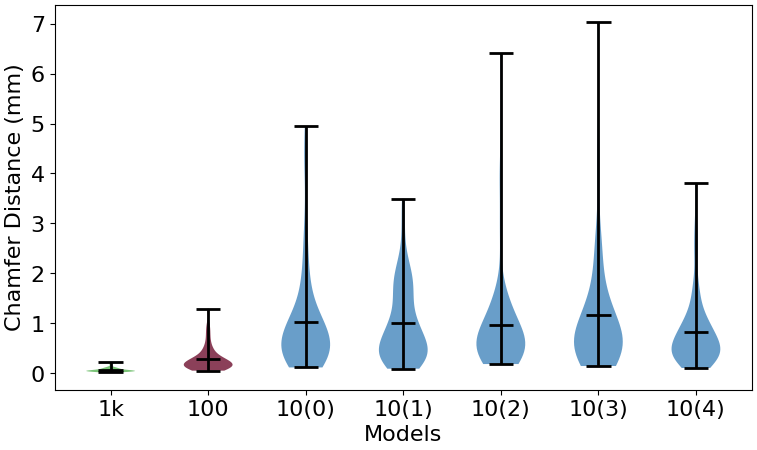}
     \caption{\textbf{Simulated tissue wrapping} - Chamfer distance between predicted and ground truth goal point clouds, on the test set.}    \label{fig:chamfer_plot_tissue_wrapping_sim}
    \vspace{-20pt}
\end{figure}

Fig.~\ref{fig:chamfer_plot_tissue_wrapping_sim} visualizes the Chamfer distance between predicted and ground truth goal point clouds on a test set of 100 unseen demonstrations.  As with retraction, there is a clear trend that more data makes the predicted goals more similar to the ground truth shapes.

\subsection{Physical Robot Experiments}
\subsubsection{Zero-Shot Sim-to-Real Transfer}

\begin{figure}[ht]
    \centering
    \includegraphics[width=1\linewidth]{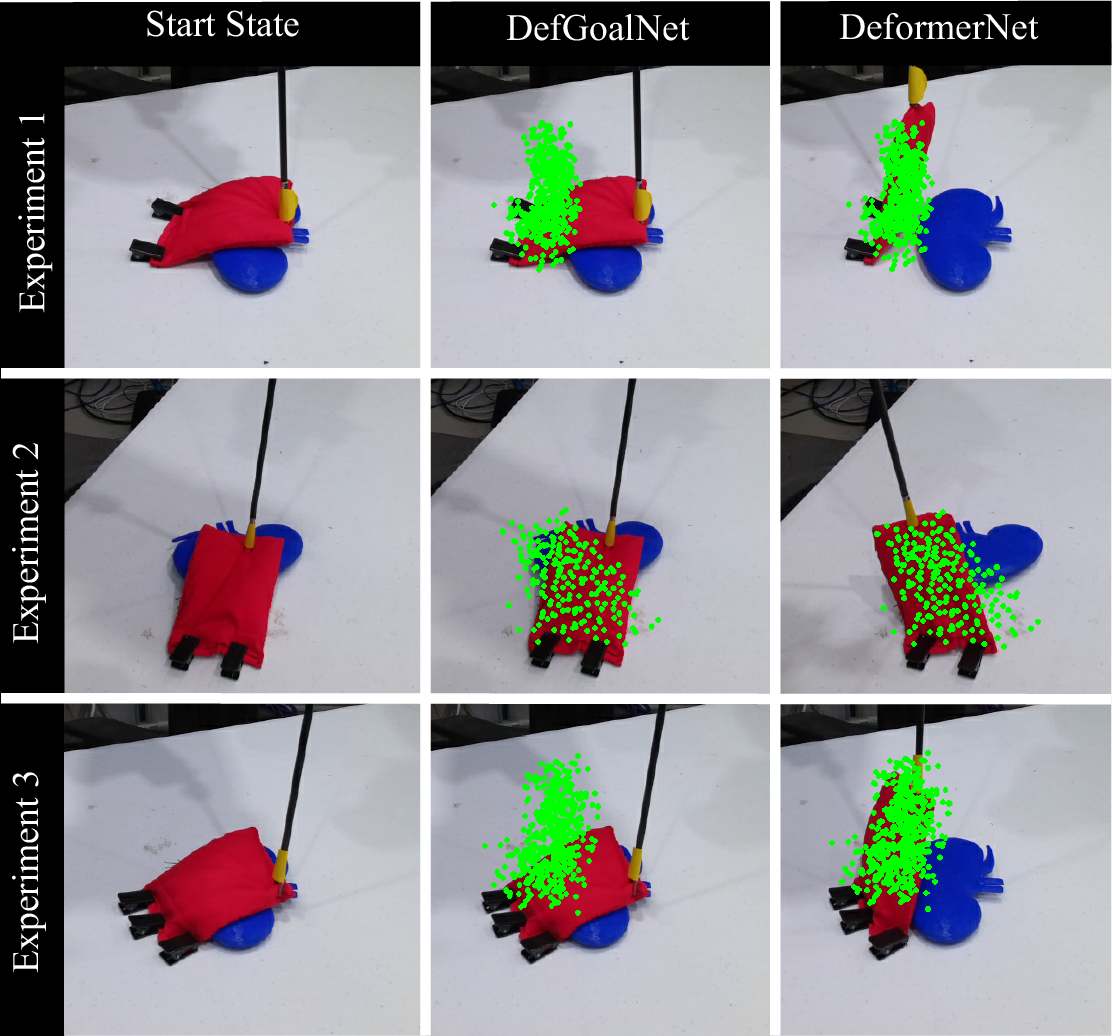}
    \caption{Sample manipulation sequences on the kidney retraction task, with the physical robot (zero-shot sim-to-real).}    
    \label{fig:sequence_retraction_kidney_real}
    \vspace{-10pt}
\end{figure}

We first evaluate whether the learned goal point clouds trained in simulation can be directly utilized to perform retraction tasks with a physical robot. We employ the model trained with 100 demonstrations in Sec.~\ref{sec:retraction_exp_sim} for this experiment.
We use a 3D-printed human-size kidney and a soft object as stand-ins for the biological kidney and deformable tissue. We conduct experiments on 3 different pose configurations of kidney and tissue. 
For each configuration, we execute the pipeline 5 times to ensure the robustness of our method. We observe that the robot always succeeds in retracting the tissue over the entire 15 runs. Figure~\ref{fig:sequence_retraction_kidney_real} illustrates representative sequences.

\subsubsection{Real Human Demos - Hand-conditioned Retraction}
\begin{figure}[ht]
\vspace{-5pt}
    \centering
    \includegraphics[width=0.8\linewidth]{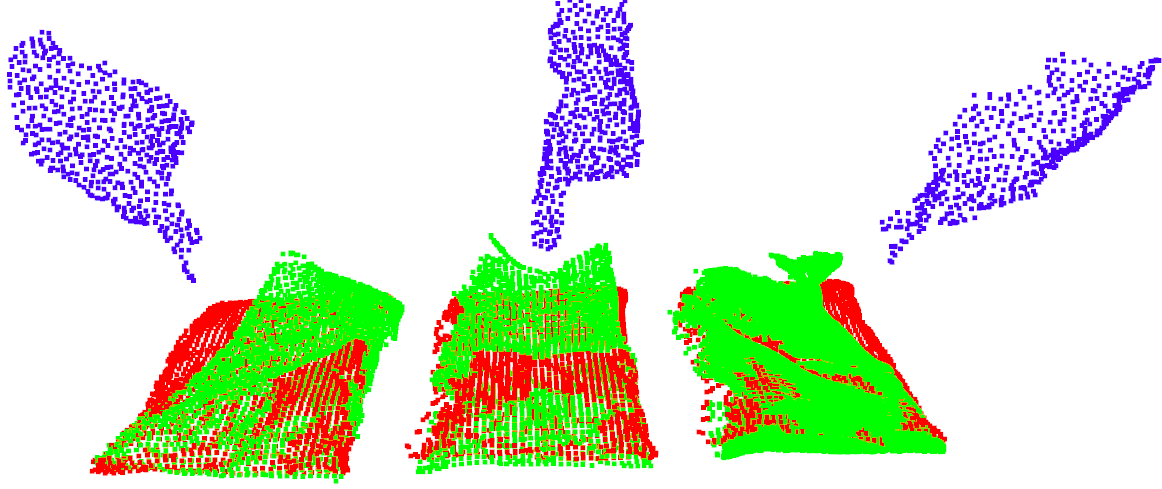} \\

    \caption{Visualization of real human demonstrations. The retraction procedure varies based on the observed hand pose. Blue: hand pose; Red: initial tissue shape; Green: human-demonstrated goal shape.}
    
    \label{fig:dataset_vis_real_retraction}
    \vspace{-10pt}

\end{figure}

We develop a mock surgical retraction task to demonstrate the learning capabilities of \ggn{} when exposed to real human demonstrations.
In surgery, a robot may assist by retracting tissue based on the location of the human surgeon's hand within the surgical scene.
Fig.~\ref{fig:dataset_vis_real_retraction} illustrates how the tissue should be retracted given the hand pose.
Motivated by this practical application, we collect a set of 20 diverse demonstrations and direct \ggn{} to learn the desired shape of the tissue with the human hand's partial-view point cloud as $\taskpcloud$ (see Fig.~\ref{fig:goalgennet_architecture}). We employ a thin deformable object as a stand-in for biological tissue to facilitate this data collection process. Our data acquisition involves manually manipulating this object using tongs, closely simulating surgical retraction. 

\begin{figure}[ht]
    \centering
    \vspace{-5pt}
    \includegraphics[width=0.24\linewidth]{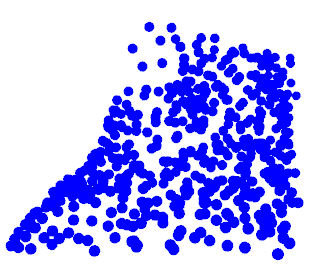} 
    \includegraphics[width=0.24\linewidth]{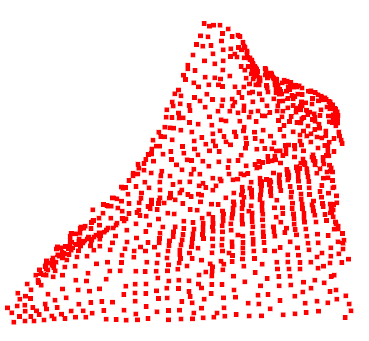}    
    \includegraphics[width=0.24\linewidth]{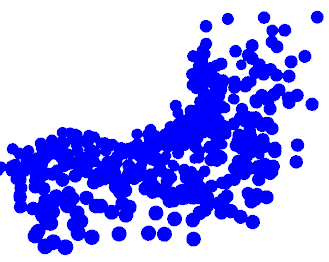} 
    \includegraphics[width=0.24\linewidth]{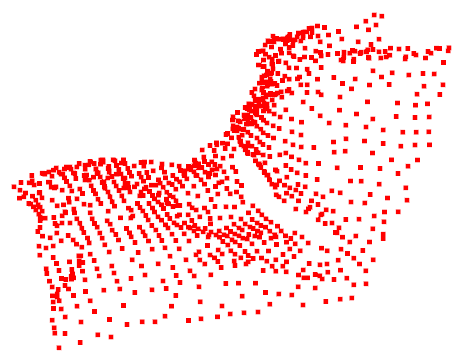} \\ 

    \caption{Predicted (blue) vs ground truth (red) test set goals.}

    \label{fig:compare_pred_gt_real_retraction}
    \vspace{-5pt}

\end{figure}

We train the model with 15 demonstrations, holding 5 demonstrations for testing purposes. The Chamfer distances between the predicted and ground truth goal point clouds on this test set are $1.72, 0.97, 1.64, 0.45,$ and $0.64$ millimeters. Fig.~\ref{fig:compare_pred_gt_real_retraction} visualizes the two worst examples of the predicted goal point clouds on the test set. The goals look similar to the ground truth, showcasing \ggn{}'s ability to generalize even when trained on a relatively small dataset. 

We evaluate the robot using three distinct goals, each predicted from distinct contextual point clouds of the human's hand pose. We execute 5 trials for each scenario to ensure robustness for a total of 15 runs. Fig.~1 visualizes a representative retraction result. Qualitatively, our method predicts goal shapes that align well with the human demonstrator's objectives and computes the necessary robot actions to drive the object to these goals.

\section{Conclusions}\label{sec:conclusions}
In this paper, we have developed a novel robotic pipeline designed for learning deformable object manipulation from demonstrations. At the heart of this pipeline lies \ggn{}, a neural network that predicts the desired object shape to successfully execute a given task. 

As for current limitations, our method assumes that the final state of the object solely defines success. In many robotic applications, the overall trajectory is a critical consideration of task success. %
To address this limitation, future research could explore methods for predicting a sequence of goal shapes instead of just a single goal instance. 
We additionally aim to broaden the application of our method, extending it to encompass domains such as real robotic surgery and robots deployed in home and warehouse environments.

We demonstrate \ggn{}'s effectiveness on a diverse set of tasks, achieving a 100\% success rate on a zero-shot sim-to-real task. Crucially, we show how contextual goal generation can be learned from relatively few demonstrations while still leveraging a control policy learned on a large and diverse dataset independent of the specific downstream task.

\clearpage\newpage
\bibliographystyle{IEEEtran}
\bibliography{bibliography}

\end{document}